\documentclass{article}
\usepackage{float}
\usepackage[utf8]{inputenc}
\usepackage{biblatex}
\bibliography{bibs.bib}  
\usepackage{PRIMEarxiv}
\usepackage{matlab-prettifier}
\usepackage[utf8]{inputenc} 
\usepackage[T1]{fontenc}    
\usepackage{hyperref}       
\usepackage{url}            
\usepackage{booktabs}       
\usepackage{amsfonts}       
\usepackage{amsmath}
\usepackage{nicefrac}       
\usepackage{microtype}      
\usepackage{lipsum}
\usepackage{fancyhdr}       
\usepackage{graphicx}       
\usepackage{xcolor}
\graphicspath{{media/}}     
\title{A MATLAB tutorial on deep feature extraction combined with chemometrics for analytical applications
\thanks{\textit{\underline{Citation}}: 
\textbf{Authors. Title. Pages.... DOI:000000/11111.}} 
}

\author{
  Puneet Mishra*, Martijntje Vollebregt \\
  Food and Biobased Research\\
  Wageningen University and Research\\
  Wageningen, The Netherlands\\
  \texttt{*puneet.mishra@wur.nl} \\
  \And
  Yizhou Ma \\
  Food Process Engineering \& Food Quality and Design\\
  Wageningen University and Research\\
  Wageningen, The Netherlands\\
  \And
  \textbf{Maria Font-i-Furnols } \\
  IRTA-Food Quality and Technology \\
  Finca Camps i Armet, 17121 Monells, Spain \\
}

\begin{document}
\maketitle

\begin{abstract}
\underline{Background}\\
In analytical chemistry, spatial information about materials is commonly captured through imaging techniques, such as traditional color cameras or with advanced hyperspectral cameras and microscopes. However, efficiently extracting and analyzing this spatial information for exploratory and predictive purposes remains a challenge, especially when using traditional chemometric methods. Recent advances in deep learning and artificial intelligence have significantly enhanced image processing capabilities, enabling the extraction of multiscale deep features that are otherwise challenging to capture with conventional image processing techniques.
Despite the wide availability of open-source deep learning models, adoption in analytical chemistry remains limited because of the absence of structured, step-by-step guidance for implementing these models. \\

\underline{Results} \\
This tutorial aims to bridge this gap by providing a step-by-step guide for applying deep learning approaches to extract spatial information from imaging data and integrating it with other data sources, such as spectral information. Importantly, the focus of this work is not on training deep learning models for image processing but on using existing open source models to extract deep features from imaging data. \\

\underline{Significance}\\
The tutorial provides MATLAB code tutorial demonstrations, showcasing the processing of imaging data from various imaging modalities commonly encountered in analytical chemistry. Readers must run the tutorial steps on their own datasets using the codes presented in this tutorial. 

\end{abstract}

\keywords{Artificial intelligence \and hyperspectral \and spectroscopy \and transfer learning \and data fusion}

\section{Introduction}
\label{sec:intro}

The exploration of the spatial distribution of chemical properties is of particular interest in analytical chemistry. Spatial distribution provides the ability to examine sample heterogeneity or even measure a large number of samples in a single scan. Such spatial insights are unattainable with conventional point measurements \cite{adams2015chemical, capitan2015recent}, which necessitate the use of imaging techniques. Although imaging techniques are often associated with RGB imaging, in analytical chemistry, a wide range of imaging methods, including multimodal techniques such as hyperspectral imaging \cite{amigo2015hyperspectral}, Raman imaging \cite{gupta2020raman}, and mass spectrometry imaging \cite{buchberger2017mass}, are utilized to explore the spatial distribution of spectral properties. Multimodal imaging typically refers to techniques capable of measuring multiple phenomena simultaneously. For example, hyperspectral imaging allows the joint measurement of both spatial and spectral properties of samples. Beyond analytical chemistry, spatially resolved spectral data hold significant value in various scientific fields such as pharmaceuticals \cite{zeng2022research}, food science and technology \cite{falcone2006imaging, liu2017hyperspectral}, and environmental science \cite{herruzo2024mass}, where understanding the localization and interaction of spectral components is critical.

A key advantage of imaging techniques in analytical chemistry is their non-destructive and non-invasive nature \cite{capitan2015recent,adams2015chemical}. Methods such as near-infrared (NIR) hyperspectral imaging \cite{amigo2015hyperspectral}, X-ray computed tomography (CT) \cite{i2009estimation}, and nuclear magnetic resonance imaging (NMRI) \cite{fan2019recent,kirtil2017recent} allow examination of internal structures and chemical compositions without damaging the sample. This capability is essential for non-destructive quality control and monitoring, where preserving samples for further testing or observing changes over time is crucial \cite{carabus2015predicting,albano2025calibration}. Advances in hardware and software have further enabled high-throughput and automated workflows \cite{capitan2015recent,adams2015chemical} using imaging systems, supporting rapid analyses in in-line scenarios.  

Spectral imaging techniques capture increasingly complex spatially distributed spectral data, and extracting meaningful information from these data has become a key research area in analytical chemistry \cite{amigo2015hyperspectral,jirayupat2021image}. Traditional image processing methods, while capable of identifying basic structural or color-based characteristics, often lack the sensitivity to capture intricate patterns in high-dimensional spectral imaging datasets \cite{debus2021deep}. Moreover, the traditional methods rely on task-specific feature extractions, which need to be manually identified and tuned for specific applications. In spectral image processing, much attention has been paid to analyzing spectral information \cite{amigo2015hyperspectral,jirayupat2021image}, while the rich spatial contextual information is rarely utilized \cite{mishra2024short,mishra2023portable,xu2022combining,xu2022complementary}. This is partly because estimating spectral information over a sample as its mean spectrum is simpler than estimating spatial features.  

With the use of pre-trained models, deep learning has revolutionized the spatial feature extraction process\cite{mishra2024short}. Pre-trained models, trained on large and diverse datasets, provide an efficient approach for extracting spatial features at various levels of abstraction \cite{saxena2020pre,han2021pre,lopes2017pre}. Architectures such as VGG \cite{Simonyan15}, ResNet \cite{he2016deep}, and EfficientNet \cite{tan2019efficientnet} can extract complex spatial patterns, ranging from edges to high-level representations of shapes, textures, and spatial relationships. These models are highly adaptable, making them suitable for various imaging modalities encountered in analytical chemistry with minimal additional training and computational resources.  

Using pre-trained models offers several advantages as they are already optimized for feature extraction, reducing the need for extensive model training and saving time and computational resources \cite{saxena2020pre,lopes2017pre}. Pre-trained models are designed to detect multiscale features \cite{han2021pre} as they provide a more comprehensive feature extraction compared to conventional approaches. Furthermore, pre-trained models are versatile and can be adapted to different imaging modalities in analytical chemistry, as demonstrated in this tutorial. These models also facilitate automated feature extraction \cite{mishra2024short,mishra2023portable,xu2022combining,xu2022complementary}, enabling users to identify meaningful patterns without requiring extensive expertise in image processing.  Fine-tuning these models \cite{mishra2021realizing,xu2022combining,xu2022complementary} further customizes them to specific imaging modalities and data types, broadening their applicability in analytical chemistry.  

This article highlights the application of pre-trained deep learning models for extracting image features in analytical chemistry. The tutorial provides MATLAB code tutorial demonstrations, showcasing the processing of imaging data from various imaging modalities commonly encountered in analytical chemistry. It demonstrates the usage of pre-trained deep learning models to extract image features from different modalities and leverage chemometric approaches for predictive modeling. Readers must run the tutorial steps on their own datasets. 

\section{Description of deep feature extraction}
\label{sec:comments}

\subsection{Key steps in deep feature extraction}
\label{sec:dl}

Deep feature extraction is a generic approach to transform raw data into meaningful features through pre-trained deep learning models. It can be applied to any type of analytical data, this tutorial focuses on the imaging modality of analytical techniques. The main motivation for considering the imaging modality is the availability of numerous pre-trained, open-source deep learning models from the field of computer vision. These models can be used to extract features from imaging modality of multi-modal analytical data without requiring additional pre-training.  For example, this tutorial demonstrates the mathematical steps for extracting image features using a pre-trained ResNet model, such as ResNet-18. ResNet has a specific architecture designed to capture multi-scale features through its residual blocks. Depending on the network architecture and the level of detail required, feature extraction can be performed at various stages of information flow within the network—for instance, after a convolution layer, a dense layer, or a normalization layer. Note the tutorial also requires ’Statistics and Machine Learning Toolbox’ for pls regression analysis.

\subsubsection{Image preparation}
\label{sec:ip}

The first step in feature extraction from images is to prepare the images with the correct dimensions and band orders. For RGB images, this process is straightforward, as most deep learning networks are trained on RGB images. However, for other modalities, such as NIR, Raman imaging, or X-ray imaging, the raw images may first need to be compressed to a limited number of bands. The number of bands depends on the input size required by the deep networks. For example, in the case of a model trained on RGB images, the input size for bands is 3, corresponding to the red, green, and blue channels. Therefore, when the network expects 3 bands, the large number of bands captured by spectroscopy techniques must be compressed into 3 dimensions. This transformation can be accomplished in several ways, such as performing a dimensionality reduction step (e.g., principal component analysis or partial least-squares regression) or using feature selection analysis. If the user has sufficient expertise in spectral peaks and their relationship to the properties of interest, manual selection of bands can also be performed. Regardless of the compression method, 3 representative bands are needed for the deep feature extraction task. 

For a more advanced approach, users can train auto-encoder-based data compression techniques \cite{tsimpouris2021using} to reduce the data to the desired number of bands. However, training an auto-encoder is computationally expensive and would require a separate tutorial article.  

Let’s assume the data has already been compressed to 3 dimensions, matching the input dimensions required by the deep network. Given an input image \( I \) with dimensions \( H \times W \times C \) (where \( H \) is the height, \( W \) is the width, and \( C = 3 \) represents RGB channels), pre-processing involves resizing and normalizing the image to meet the ResNet model’s input requirements.  

1. Resizing: Resize the input image to \(224 \times 224 \times 3\) (for ResNet-18). Note this resizing is ResNet-18, depending on the network used the user must check the input dimension to prepare their data for feature extraction. There are different ways to resize images and in the following part, resizing images in MATLAB will be demonstrated.
   
2. Normalization: Normalize each pixel by subtracting the mean and dividing by the standard deviation (based on ImageNet statistics):
   \[
   I' = \frac{I - \mu}{\sigma}
   \]
   where \( \mu \) and \( \sigma \) are the mean and standard deviation vectors for each color channel (RGB), respectively.\\

Pre-processing with resizing and normalization makes sure the samples matches the requirement for input to the deep network.

Before understanding the data processing with deep layers it is of important to understand the function of each layer in the resnet model. Here different layers are described.\\

\textbf{Convolutional layers}\\
Convolutional layers slide (convolve) small filters across the image (or feature map) to detect features like edges, corners, or more complex textures. Early layers detect simple things; deeper layers detect more abstract patterns (like faces, objects). Mathematically, it can be approximated as the weighted sum over a small region of the input.\\

\textbf{Batch Normalization Layers} \\
After convolution, the output is normalized (zero mean, unit variance). This reduces "internal covariate shift" (changes in distribution during training), so training is faster and more stable. Similar to smoothing out the activations so the network learns better.\\

\textbf{ Activation Layers (ReLU)}\\
ReLU (Rectified Linear Unit) is a simple mathematical activation function f(x) = max(0, x). It keeps positive values and sets negative ones to zero. This non-linearity allows the network to learn very complex, non-linear mappings from input to output.\\

\textbf{Residual (Skip) Connections}\\
Instead of learning a direct mapping, ResNet learns a "residual" — the difference between input and output. A skip connection adds the input x directly to the output of some layers. This helps gradients flow better during backpropagation and allows very deep networks (like 50, 101, 152 layers) to be trained.\\

\textbf{Pooling Layers} \\
Usually Max Pooling layers takes the maximum value from small patches to reduce the spatial size (width and height) of feature maps while keeping important information. Makes the network faster and more focused on high-value features.\\

\textbf{Global Average Pooling (GAP)}\\
Instead of flattening a giant feature map, ResNet usually applies GAP. GAP takes each feature map and averages it down to a single value this reduces parameters a lot and avoids overfitting.\\

\textbf{Fully Connected (Dense) Layers}\\
At the very end, after all the convolutional and pooling layers. Dense layer relates the high-level features and output to make the final predictions.\\

\subsubsection{Initial Convolution and Pooling Layers}
\label{sec:icp}

The preprocessed image \( I' \) is passed through the initial layers, which include a convolutional layer, batch normalization, ReLU activation, and max pooling. At every step, sequential transformation of data takes place. Features can be taken out from pooling layers.

1. Convolution:
   \[
   F_1 = I' * K
   \]
   where \( * \) denotes the convolution operation, and \( K \) is the convolutional kernel.

2. Batch Normalization:
   \[
   F_1' = \text{BatchNorm}(F_1)
   \]

3. ReLU Activation:
   \[
   F_1'' = \text{ReLU}(F_1')
   \]

4. Max Pooling:
   \[
   F_{\text{pool}} = \text{MaxPool}(F_1'')
   \]

\subsubsection{Residual Blocks (Feature Extraction)}
\label{sec:rb}

ResNet’s core is composed of residual blocks, each containing convolutional layers with a skip (identity) connection that helps avoid vanishing gradients. In a residual block, we apply:

1. First Convolutional Layer:
   \[
   F_2 = F_{\text{in}} * K_1
   \]

2. Batch Normalization and ReLU Activation:
   \[
   F_2' = \text{ReLU}(\text{BatchNorm}(F_2))
   \]

3. Second Convolutional Layer:
   \[
   F_3 = F_2' * K_2
   \]

4. Batch Normalization:
   \[
   F_3' = \text{BatchNorm}(F_3)
   \]

5. Skip Connection:
   \[
   F_{\text{res}} = F_3' + F_{\text{in}}
   \]

6. Final ReLU Activation:
   \[
   F_{\text{out}} = \text{ReLU}(F_{\text{res}})
   \]

The above steps are repeated for each residual block in ResNet-18, progressively capturing more complex features.

\subsubsection{Global Average Pooling}
\label{sec:gp}

After the last residual block, global average pooling (GAP) is applied to condense each feature map into a single average value. Let \( F_{\text{last}} \) be the output feature map from the final residual block. GAP is calculated as:

\[
F_{\text{GAP}} = \frac{1}{h \times w} \sum_{i=1}^{h} \sum_{j=1}^{w} F_{\text{last}}[i, j]
\]

This operation reduces \( F_{\text{last}} \) to a 1D vector, where each element represents an average over the spatial dimensions. The output from the GAP layer is the feature vector, denoted as:

\[
\text{Feature Vector} = F_{\text{GAP}}
\]

For ResNet-18, this feature vector has a dimensionality of 512, which corresponds to the number of filters in the last convolution layer. The feature vector extracted from deep networks can be considered as a spectrum which also posses several challenges similar to multivariate data such as multi-collinearity. Chemometrics-based latent space techniques allows de-tangling the multi-collinearity for such signals. The extracted feature vector is then ready for downstream tasks, such as classification, clustering, and prediction/regression. In case of multi-modal data such as imaging together with spectroscopy, the extracted feature can be fused together with spectral data using multi-block fusion approaches. A case of multi-block fusion will also be demonstrated.

\section{Hands on tutorial}
\label{sec:results}

\subsection{RGB images of plant based meat analogue for predicting fibrousness}
\label{sec:pb2}

The dataset \cite{ma2024quantitative} consists of 82 close-up RGB images of various plant-based materials developed as alternatives to musle meat. These images were captured under controlled lighting conditions in a photo booth using a SONY A6000 digital camera (Japan) equipped with a 100mm macro lens (Tokina, Tokyo, Japan). The camera was positioned approximately 20 cm from each sample. The images contain structural information that describes the fibrousness of the meat analogue, which is a key quality attribute to resemble the fibrous texture of cooked muscle meat. The images were rated by domain experts for their fibrousness, which is the target values for prediction in this application.  

Specifically, a quantitative visual assessment panel was conducted through an online survey created in Qualtrics (Washington, USA). For this study, 26 experts with experience evaluating the macrostructures of meat analogues assessed fibrousness using a visual analog scale ranging from 0 to 100. The scale included five labeled points: not fibrous, somewhat fibrous, fibrous, very fibrous, and extremely fibrous, corresponding to intervals of 0, 25, 50, 75, and 100. Additionally, images of a non-fibrous sample and an extremely fibrous sample were provided at the start of each question as reference points. For each sample, one of five images was randomly selected and presented to an expert, ensuring that each image received at least five evaluations. The median of these evaluations was used as the expert rating for visual fibrousness.  

\begin{figure}[H]
  \centering
  \includegraphics[width=\textwidth]{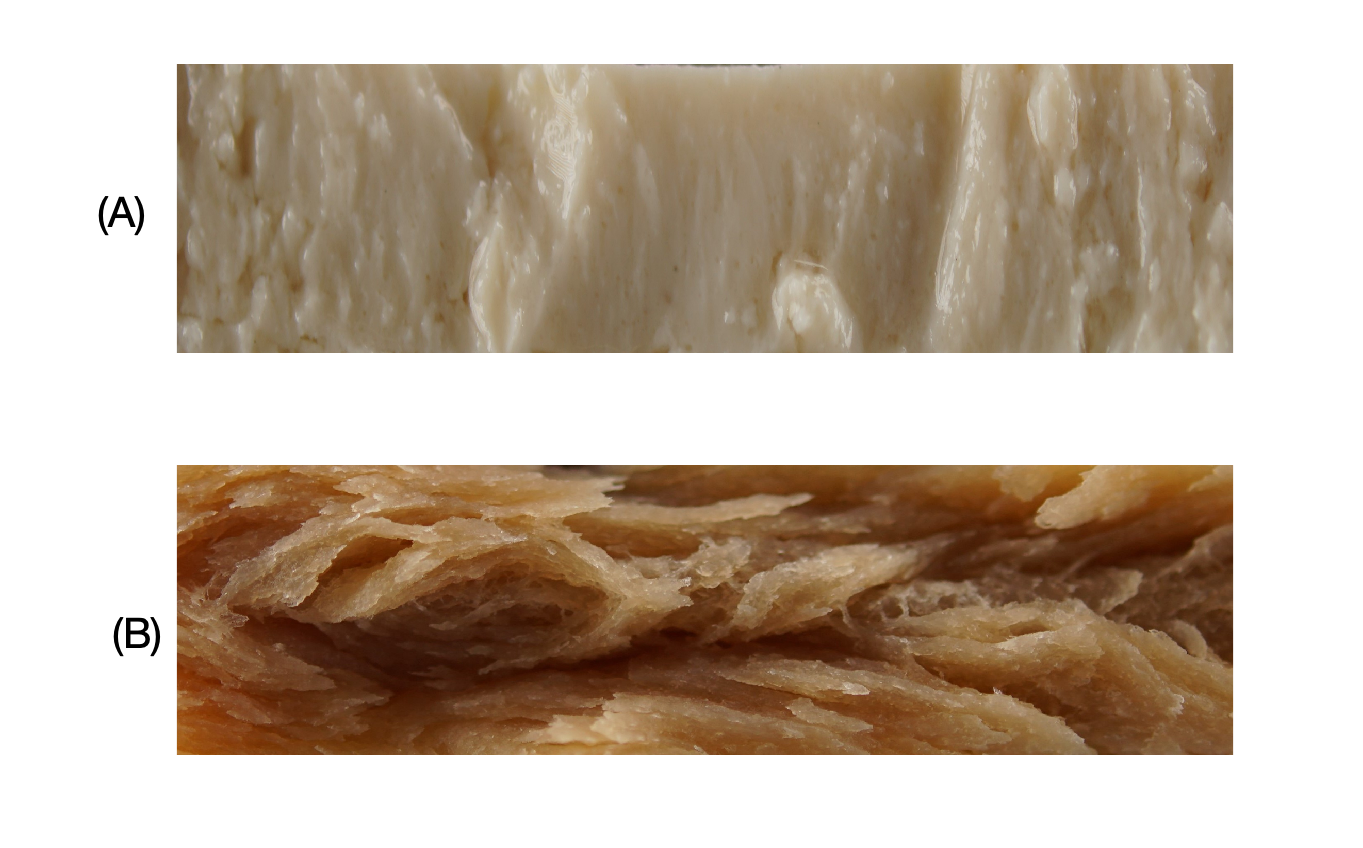}
  \caption{Sample RGB images of extruded plant based materials. }
  \label{fig:fig1}
\end{figure}

An example of the images can be seen in Fig. \ref{fig:fig1}. In this figure, the samples differ in color, texture, and fibrousness. While color is relatively easy to extract, texture and fibrousness features are more challenging to capture directly and have traditionally required several custom operations \cite{ma2024quantitative}.  

Fig. \ref{fig:fig1} was plotted using the following code.

\begin{lstlisting}[style=Matlab-editor]
%% Sample Matlab code
subplot(2,1,1)
imagesc(image1);axis off;
subplot(2,1,2);
imagesc(image2);axis off;
\end{lstlisting}

Deep feature extraction can be performed in a straightforward manner, as demonstrated below. In this example, an existing ResNet-18 deep learning model is loaded into the MATLAB workspace, and the input size is printed. The input size is crucial because all sample images must match the network's input dimensions to pass through it successfully.  

For ResNet-18, the input size is \(224 \times 224 \times 3\), where the first two dimensions represent spatial size (height and width), and the third dimension represents the spectral channels. In the case of RGB images, the three spectral channels correspond to the red, green, and blue colors.  

The model can be loaded using the following code. Note that this code is specific to ResNet-18, but several other pre-trained models are available in MATLAB and can be loaded using a similar command. To load a different pre-trained model, users must specify the correct network name. Some common alternatives include 'densenet201', 'resnet50', 'resnet101', 'darknet19', 'vgg16', and 'vgg19'.

\begin{lstlisting}[style=Matlab-editor]
%% Sample Matlab code
net = resnet18;
inputSize = net.Layers(1).InputSize;
\end{lstlisting}

Once the pre-trained model is loaded, it will appear in the MATLAB workspace as `net`. The next step is to transform all the images to match the network's input size. After resizing, the images need to be stored in a datastore compatible with the deep network. The resizing step in MATLAB can be performed using the following code.  

\begin{lstlisting}[style=Matlab-editor]
%% Sample Matlab code
im = imresize(imread(imagefiles),[224 224]);
augimdsTrain = augmentedImageDatastore(inputSize(1:2),im);
\end{lstlisting}

Once the images are resized to match the network's input dimensions, the next step is to perform a forward pass of all the images through the deep learning model. The feature information is extracted just before the network's output layer. For ResNet-18, this corresponds to the `pool5` layer.  

In the following code, the feature vector is extracted after the `pool5` layer. Users can extract information from any layer in the network. Changing the location of feature extraction simply requires modifying the layer parameter in the code. For example, setting `layer = 'pool5'` will extract the feature vector from the `pool5` layer, while setting `layer = 'pool4'` will extract the features from the `pool4` layer.  

\begin{lstlisting}[style=Matlab-editor]
%% Sample Matlab code
layer = 'pool5';
featuresTrain = activations(net,augimdsTrain,layer,'OutputAs','rows');
\end{lstlisting}

Once the feature vector is extracted, all traditional analyses, such as plotting for visualization or supervised/unsupervised analysis, can be performed. If needed, dimensionality reduction techniques can be applied to visualize any grouping within the data. This can be done using straightforward methods like principal component analysis (PCA) or more advanced nonlinear techniques such as t-distributed stochastic neighbor embedding (t-SNE).  

In Fig. \ref{fig:fig2}, a feature vector corresponding to a single image is plotted. As observed, the feature vector is multivariate with 512 dimensions. The main challenge in interpreting the feature vector is the lack of x-axis labels. In typical analytical signals, the x-axis often represents a known quantity with physical meanings, such as wavelengths or wave numbers in optical spectroscopy. However, in the case of deep features, it is unclear what each variable represents. While all variables capture spatial information at different scales, their exact definitions remain ambiguous. Though with interpretability challenges, the extracted feature vector can still be analyzed using the multivariate analysis in chemometrics. 

The characteristic vector shown in Fig. \ref{fig:fig2} is plotted using the following code:

\begin{lstlisting}[style=Matlab-editor]
%% Sample Matlab code
plot(featuresTrain');
xlabel('Deep features');ylabel('Intensity');
\end{lstlisting}

\begin{figure}[H]
  \centering
  \includegraphics[width=\textwidth]{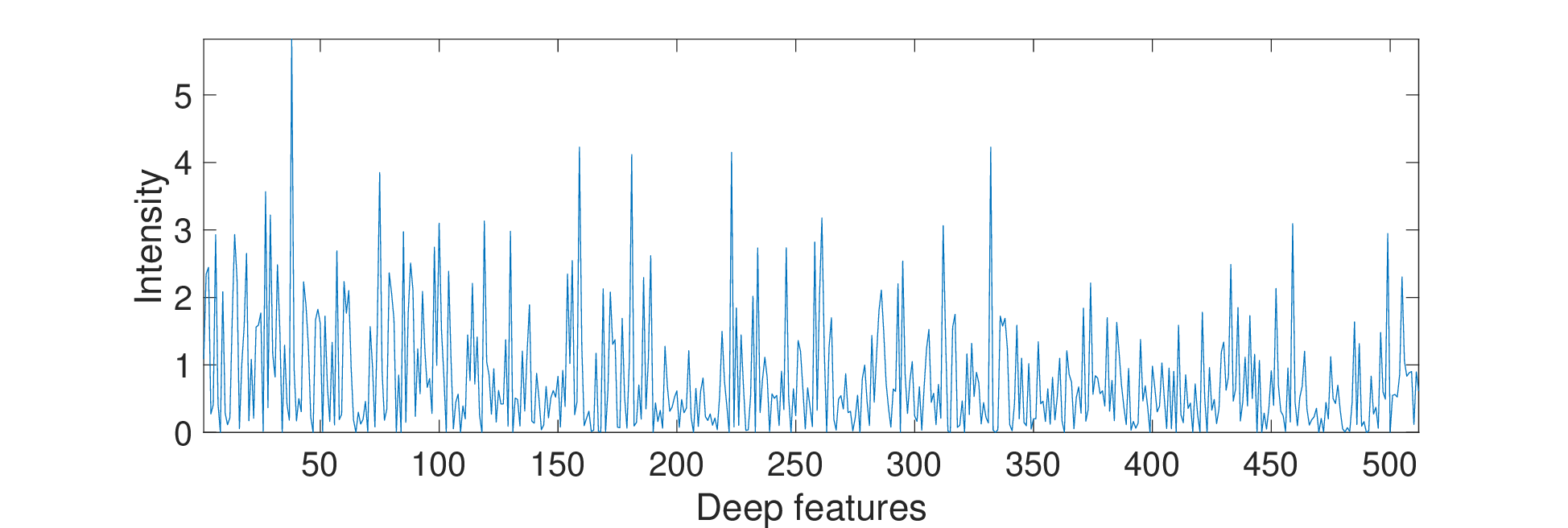}
  \caption{Feature vector extracted form an image using the 'resnet18' network. }
  \label{fig:fig2}
\end{figure}

Once the deep features are extracted from all images, they can be analyzed and used for predictive modeling. In this case, the feature vectors are used to perform partial least-squares (PLS) regression with the median fibrousness scores provided by the expert panel. PLS is particularly useful for analyzing feature vectors because they often contain a large number of variables, many of which are highly correlated. PLS addresses both issues through data compression.  

The PLS analysis of the feature vectors and the response variables can be executed using the following code. The PLS model also requires optimization of the optimal number of dimensions for data compression. This optimization can be performed through cross-validation, as demonstrated in the code below. In the provided code, `x` and `y` represent the calibration set, while `xt` and `yt` are the test set. Once the model is optimized and trained on `x` and `y`, it can be applied to `xt` and `yt`.  

\begin{figure}[H]
  \centering
  \includegraphics[width=\textwidth]{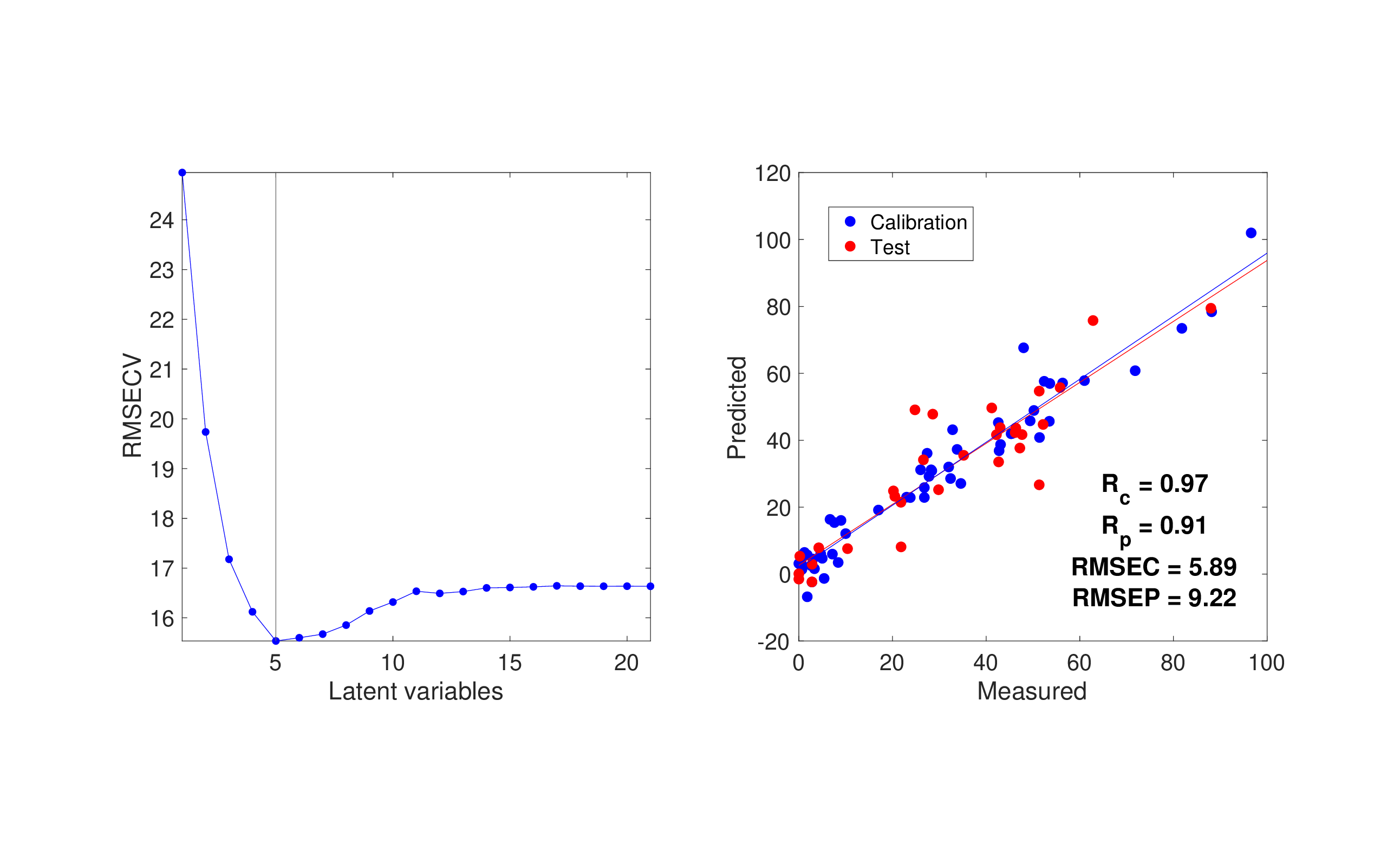}
  \caption{Cross-validation and prediction plot for PLS regression analysis on deep image features.}
  \label{fig:fig3}
\end{figure}

Running the code generates a cross-validation plot, shown in Fig. \ref{fig:fig3}, where the root mean squared error of cross-validation (RMSECV) is plotted as a function of latent variables. The minimum point on the plot indicates the optimal number of latent variables to use.

\begin{lstlisting}[style=Matlab-editor]
%% Sample Matlab code
x = features;
y = responses;
subplot(1,2,1)
[XL,YL,XS,YS,beta,PCTVAR,MSE,stats] = plsregress(x,y,20,'CV',52);
plot(sqrt(MSE(2,:)),'.-b','MarkerSize',20);set(gcf, 'Color' , 'w' );xlabel('Latent variables');ylabel('RMSECV');
[~,ind] = min(sqrt(MSE(2,:)));
xline(ind);
\end{lstlisting}

Once the optimal model parameters are identified, the final model can be constructed. In this case, five latent variables were used to calibrate the final model. The following code executes the development of the optimal model and tests it on the test set. The code execution produces a prediction plot shown in Fig. \ref{fig:fig3}.  

As observed, the deep features achieved a high correlation (>0.90) to the fibrousness in both the calibration and test sets. The prediction error (RMSEP) is less than 10 points on a 100 point scale based on the expert panel. The prediction error indicate that the PLS model can estimate fibrousness with practical significance and guide the production and optimization of plant-based meat analogue.

\begin{lstlisting}[style=Matlab-editor]
%% Sample Matlab code
subplot(1,2,2)
[XL,YL,XS,YS,beta,PCTVAR,MSE,stats] = plsregress(x,y,ind);
predsc = [ones(size(y,1),1) x]*beta;
predsp = [ones(size(yt,1),1) xt]*beta;
plot(y,predsc,'.b','MarkerSize',30); hold on
plot(yt,predsp,'.r','MarkerSize',30);
lsline;
legend('Calibration','Test');
xlabel('Measured');ylabel('Predicted');
[rmc,rmsc] = rmse(y,predsc);
[rmp,rmsp] = rmse(yt,predsp);
title(['R_c = ' num2str(round(rmc,2)) ' R_p = ' num2str(round(rmp,2)) ' RMSEC = '  num2str(round(rmsc,2)) ' RMSEP = ' num2str(round(rmsp,2))]);
\end{lstlisting}

\subsection{X-ray CT images of meat}
\label{sec:xray2}

\begin{figure}[H]
  \centering
  \includegraphics[width=\textwidth]{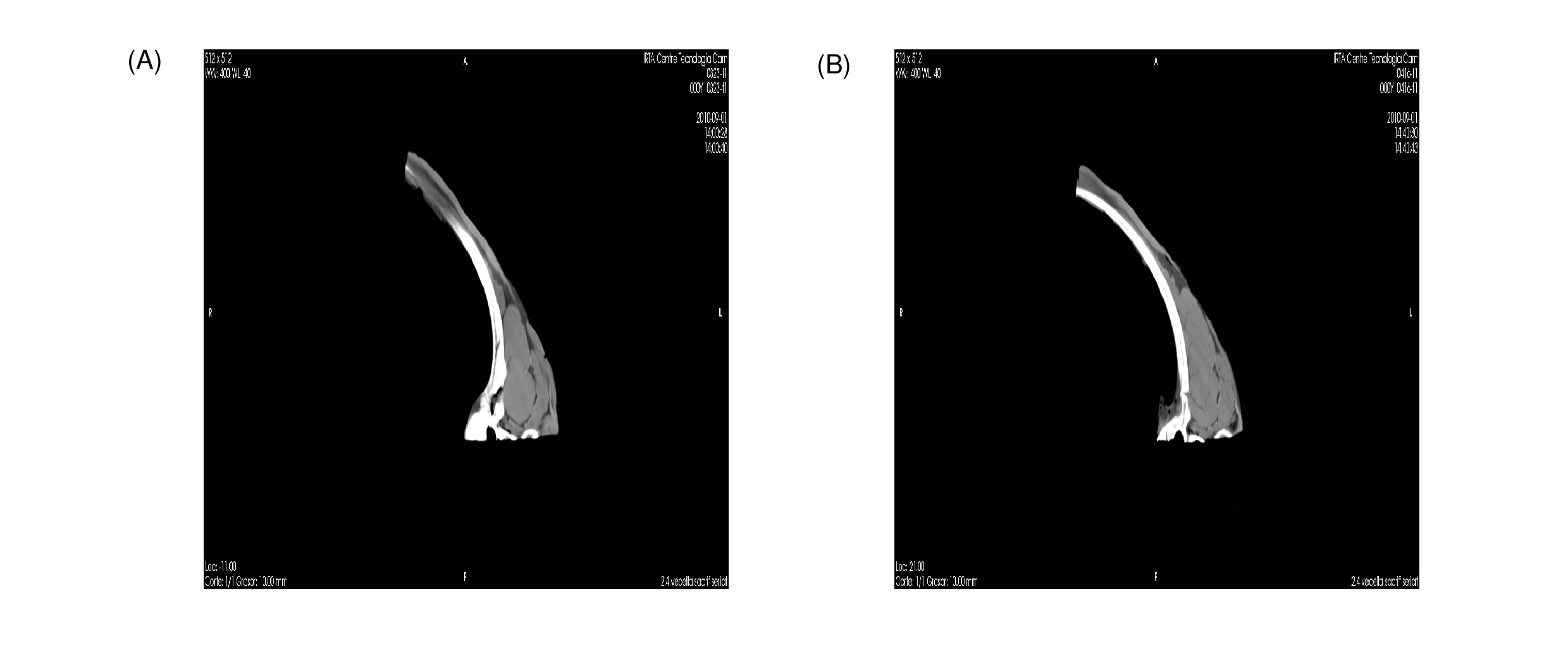}
  \caption{Example CT images of the rib chops of beef.}
  \label{fig:fig8}
\end{figure}

The second example is of X-ray CT image processing, where the goal is to predict the fat content in beef rib chops using X-ray CT images \cite{font2014composition}. The purpose of using deep feature extraction is to directly input an X-ray slice, instead of the complete CT data, to predict the fat content. In practice, generating CT data is more time- and power-consuming, and handling such data is complex. Making predictions solely based on an X-ray projection is much faster, consumes less power, and is more practical, as X-ray images are much lower in cost compared to CT scanners. To demonstrate the potential use of processing such data, a dataset containing 180 images of beef rib chops was used. Each sample was first scanned with the CT scanner and then manually dissected to measure the amount of fat in each sample. An example of CT images (but only 2D X-ray projections, as CT is a 3D volume and cannot be plotted in text) can be seen in Fig. \ref{fig:fig8}. The two CT images show differences in gray-level contrasts. Differences in hue levels in the same meat samples indicate the presence of meat, fat and bones, as noted in \cite{i2009estimation}. The white, curby region in the image corresponds to the rib bone of the chop, the light grey to the lean and the dark grey to the fat. The images in the subplot can be plotted using the following command. A typical approach to processing CT data of meat involves extracting the Hounsfield values from the data and then running regression on these data. However, such a method requires the entire CT dataset to estimate Hounsfield values. In this study, out of the 180 samples, every third measurement was used as a test set, and the remaining measurements were used for model optimization and calibration.

\begin{lstlisting}[style=Matlab-editor]
%% Sample Matlab code
subplot(1,2,1)
imagesc(image1);axis off;
subplot(1,2,2);
imagesc(image2);axis off;
\end{lstlisting}

Extraction of deep features from CT images is also straightforward, similar to RGB images. For example, the images in Fig. \ref{fig:fig8} can be directly resized to the model input size and fed into the deep network to extract features. A deep model can be loaded with the following command.

\begin{lstlisting}[style=Matlab-editor]
%% Sample Matlab code
net = resnet18;
inputSize = net.Layers(1).InputSize;
\end{lstlisting}

Secondly, resize the images to be used as input to the deep model with the following command.

\begin{lstlisting}[style=Matlab-editor]
%% Sample Matlab code
im = imresize(imread(imagefiles),[224 224]);
augimdsTrain = augmentedImageDatastore(inputSize(1:2),im);
\end{lstlisting}

Thirdly, pass the images through the network to extract deep features with the following command.

\begin{lstlisting}[style=Matlab-editor]
%% Sample Matlab code
layer = 'pool5';
featuresTrain = activations(net,augimdsTrain,layer,'OutputAs','rows');
\end{lstlisting}

An example of the deep features extracted from X-ray images and Hounsfield values from CT data is shown in Fig. \ref{fig:fig9}. The plots can be generated with the following code. In the Hounsfield value plot (Fig. \ref{fig:fig9}A), each peak corresponds to either fat, muscle, or bone. The regions between peaks represent mixed voxels in the CT images. In the case of deep features, several peaks can be observed, but interpreting them is challenging, as they are generated by a pre-trained deep network.

\begin{lstlisting}[style=Matlab-editor]
%% Sample Matlab code
subplot(1,2,1)
plot(Hounsfield_spectra(1,:));
xlabel('Hounsfield values');
ylabel('Frequency');
subplot(1,2,2)
plot(deep_features(1,:));
xlabel('Deep features');
label('Intensity');
gtext('(A)');
gtext('(B)');
\end{lstlisting}

\begin{figure}[H]
  \centering
  \includegraphics[width=\textwidth]{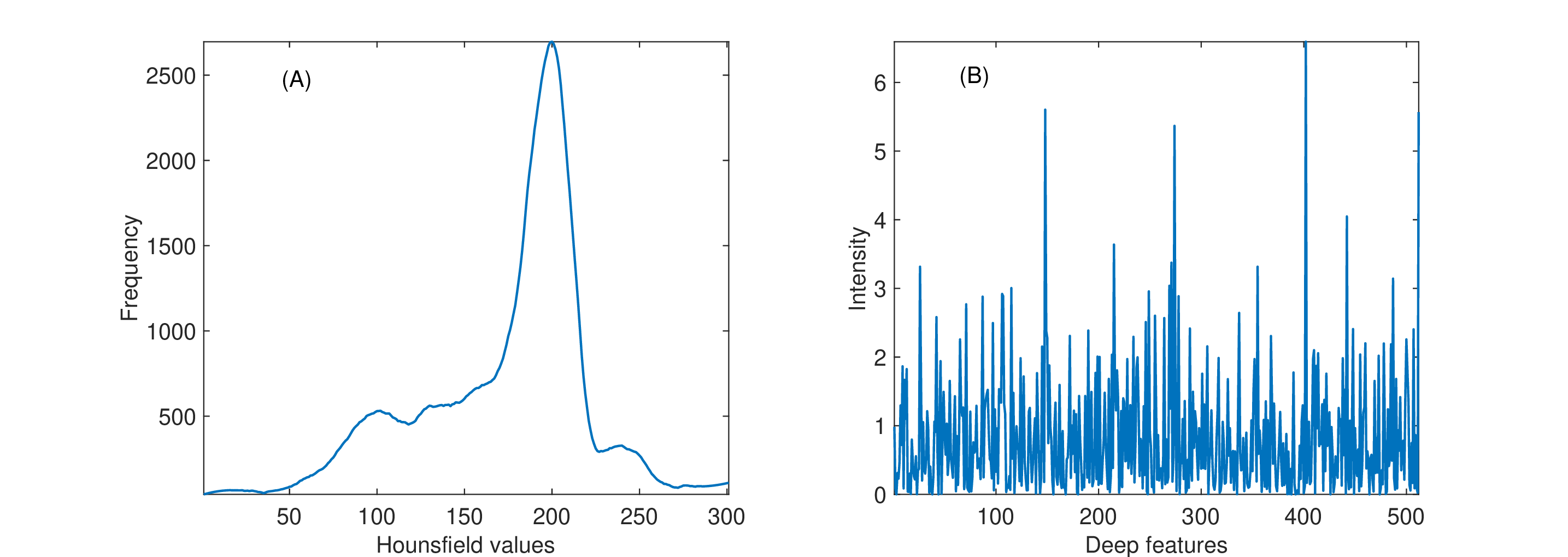}
  \caption{Example Hounsfield values of a CT image sample (A) and deep feaures from a 2D X-ray projection (B).}
  \label{fig:fig10}
\end{figure}

Models can be trained on both the deep features and the Hounsfield values to predict the fat content. Since both data sets exhibit high multicollinearity, PLS regression is an optimal choice for such data. The following code runs a PLS regression optimization to find the optimal number of latent variables. The code provides a cross-validation plot, where the optimal number of latent variables can be determined based on the minimum of the cross-validation errors, as shown in Fig. \ref{fig:fig10}A. Cross-validation analysis was performed using leave-one-out cross-validation. As can be seen in Fig. \ref{fig:fig10}A, both the deep features and Hounsfield values explain the fat content in rib cuts well. The cross-validation error decreased for both the deep features and Hounsfield values, and after reaching the minimum, it began to increase. The optimal number of latent variables was determined based on the minimum of the cross-validation error plots in Fig. \ref{fig:fig10}A.

\begin{lstlisting}[style=Matlab-editor]
[~,~,~,~,~,~,MSE] = plsregress(spectra,response,50,'CV',size(response,1));
plot(sqrt(MSE(2,:)),'.-b','Markersize',20,'LineWidth',2);xlabel('Latent variables');ylabel('RMSECV'); 
axis tight;axis square;
[~,~,~,~,beta] = plsregress(spectra,response,optimal_components);
\end{lstlisting}

The optimal models, based on cross-validation analysis, can be used to predict fat values on the test set using the following code. The prediction results for the meat data are shown in Fig. \ref{fig:fig10}B. Both the deep features and Hounsfield values were able to predict fat content (in grams) with errors of approximately 196 and 130, respectively. Hounsfield value-based predictions had lower errors than those using deep features. This was expected, as Hounsfield values capture 3D information, while the deep features are based only on a 2D X-ray projection, missing information about the thickness of the meat slice. If depth information of the meat slice could be combined with the 2D X-ray projection, fat prediction could be enhanced. In practice, this would mean using both a 2D X-ray camera and a 3D depth camera to measure the meat sample simultaneously. However, exploring such technology is beyond the scope of this tutorial.

\begin{lstlisting}[style=Matlab-editor]
predictions = [ones(size(test_set,1),1) spectra_test]*beta;
plot(yt,predss,'.b','MarkerSize',30);
xlabel('Measured fat');
ylabel('Predicted fat');
lsline;
[~,rms]=rmse(response_test,predictions);
title(['RMSE = ' num2str(round(rms,2))]);
\end{lstlisting}

\begin{figure}[H]
  \centering
  \includegraphics[width=\textwidth]{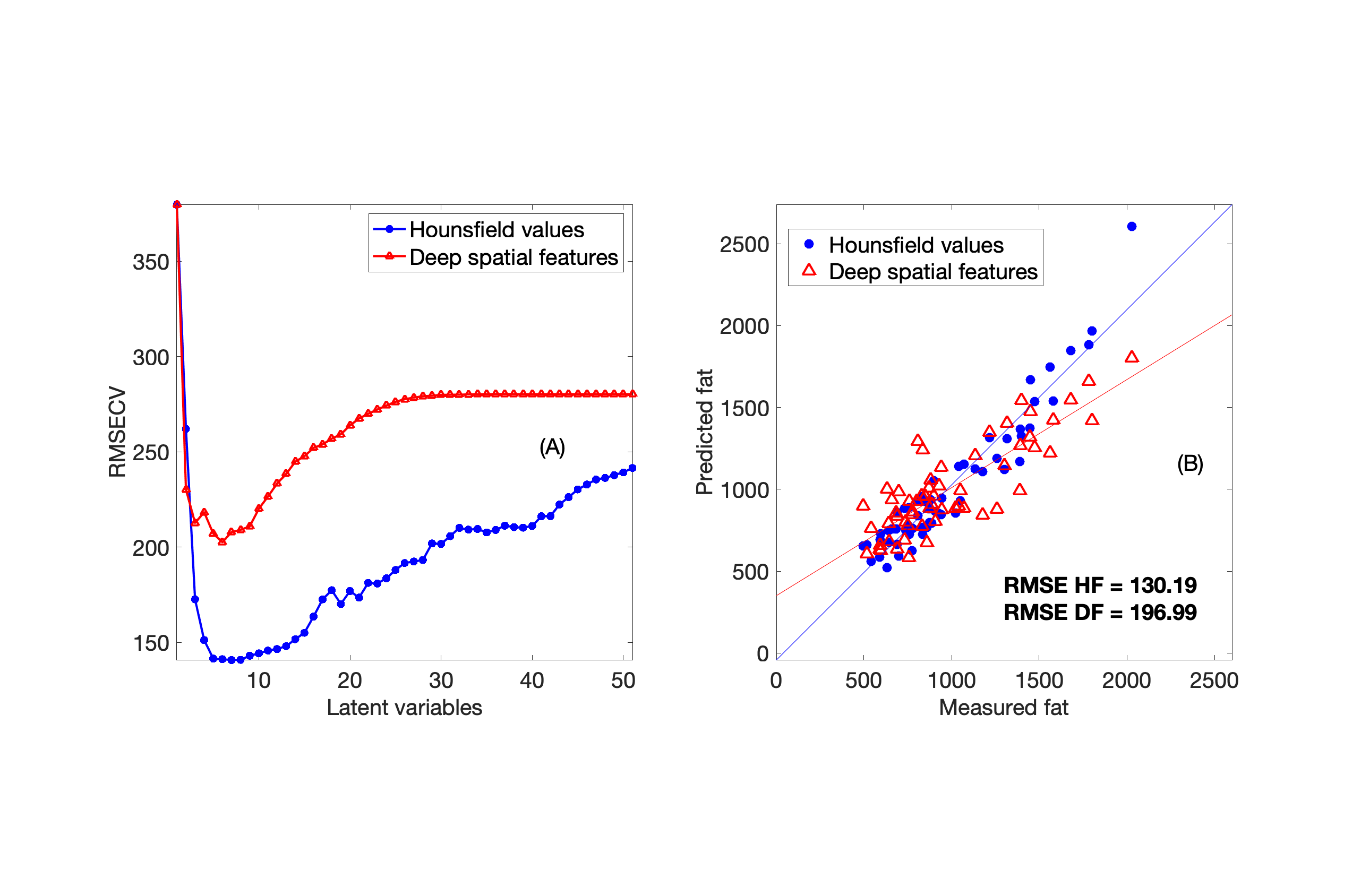}
  \caption{Cross-validation results (A) and prediction plot (B) for analysis meat CT data.}
  \label{fig:fig9}
\end{figure}

\subsection{Hyperspectral images of pork belly fat for predicting fat hardness}
\label{sec:hsi2}

The third example concerns hyperspectral imaging (HSI) of pork belly fat to predict fat hardness \cite{albano2025visible, mishra2024x}. Pork is a popular meat product, especially pork belly, where maintaining high quality throughout production is essential. Fat characteristics, particularly firmness, play a critical role in determining pork quality, as firmness is strongly influenced by fatty acid composition. This is significant because unsaturated fatty acids tend to be softer and are more prone to oxidation, affecting both texture and stability. Traditional methods for assessing fat firmness, such as shear force measurement, are often destructive, time-consuming, and labor-intensive. Consequently, there is strong industry interest in developing rapid, non-invasive sensors for evaluating fat firmness.\\

HSI data consists of two modalities: spatial and spectral. The spatial modality captures structural information, while the spectral modality provides chemical composition data. Fat composition is characterized by a complex matrix of structural features that impact both mechanical properties (such as firmness) and compositional qualities. The spectral information primarily reflects chemical characteristics, whereas spatial features capture physical attributes such as shape, texture, and surface structure.\\

The HSI dataset includes 182 pork belly samples randomly selected from the left side of half-carcasses, 24 hours post-mortem, at commercial pig slaughterhouses in Spain. Imaging was conducted using a line-scan hyperspectral camera (Spectronon, Resonon Inc., Bozeman, MT, USA), mounted at a 50 cm height on a conveyor platform. The conveyor, driven by a stepper motor, moved at a speed of 0.03 m/s. Four 50 W tungsten-halogen lamps illuminated the field of view, and images were captured in reflectance mode across wavelengths from 386 to 1015 nm \cite{albano2025visible}.\\

Belly firmness was measured using finger pressure test according to a predetermined scale: "(1) firm fat, no finger mark, no floppy; (2) firm fat, no finger mark, partly floppy; (3) soft fat, finger mark remains, floppy; (4) soft fat, finger mark remains, very floppy; (5) soft fat, finger mark remains, very floppy, oily”. The finger pressure was assessed by two trained evaluators, and the average of the scores was used, with lower scores indicating greater firmness \cite{albano2024pork}.\\

In close-range HSI, most analyses traditionally rely on spectral data alone, treating HSI primarily as a point spectrometer for gathering localized spectra. However, recent studies are increasingly utilizing spatial data to improve model accuracy and assess spatial consistency or variability in sample imaging. Some studies have enhanced model performance by manually extracting spatial features, such as gray-level co-occurrence matrices, from pseudo-color representations of spectral images. Although these combined approaches show promise, they are limited by the complexity and manual tuning required for feature extraction. Conversely, deep learning has emerged as a powerful tool to model spatial and spectral data in hyperspectral imaging.\\

\begin{figure}[H]
  \centering
  \includegraphics[width=\textwidth]{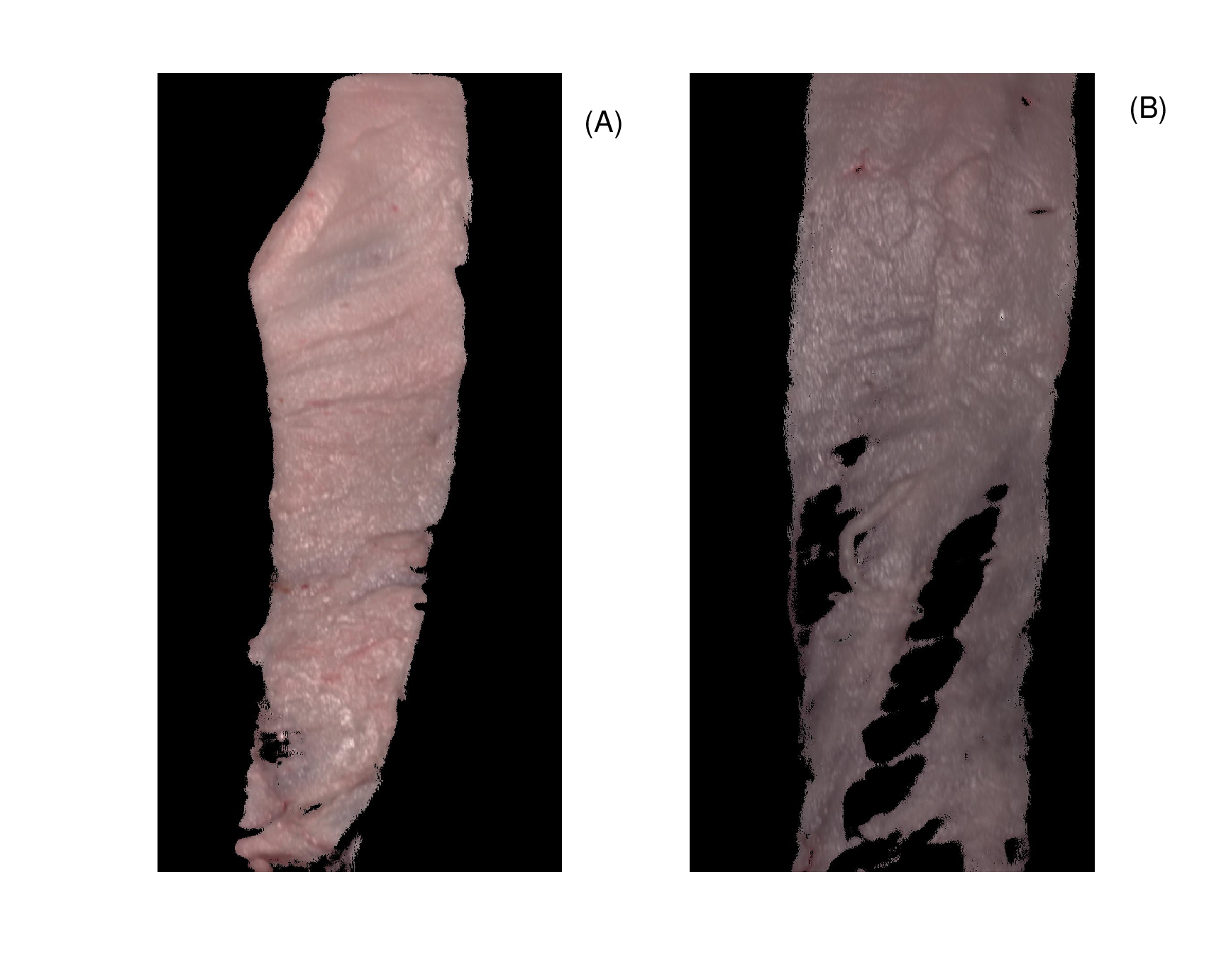}
  \caption{Psuedo segmented RGB images of pork belly to be used for deep feature extraction.}
  \label{fig:fig4}
\end{figure}

In this study, previously segmented data was used as described in \cite{albano2025visible}. Fat samples were segmented to remove background interference. Mean spectra for each sample were also computed from the segmented data. The pseudo-color images were constructed using bands 750, 670, and 500 as red, green, and blue. Some example pseudo images are shown in Fig. \ref{fig:fig4}. In the case of pork belly, it was possible to generate the pseudo RGB images of belly fat, as the camera also captures the visible bands. If spectral imaging is performed only in the near-infrared range, the user must find the optimal combination of pseudo-RGBs that provide high contrast to the physical features. Belly images in Fig. \ref{fig:fig4} are plotted using the following code: \\

\begin{lstlisting}[style=Matlab-editor]
%% Sample Matlab code
subplot(1,2,1)
imagesc(image1);axis off;
subplot(1,2,2);
imagesc(image2);axis off;
\end{lstlisting}

\begin{figure}[H]
  \centering
  \includegraphics[width=\textwidth]{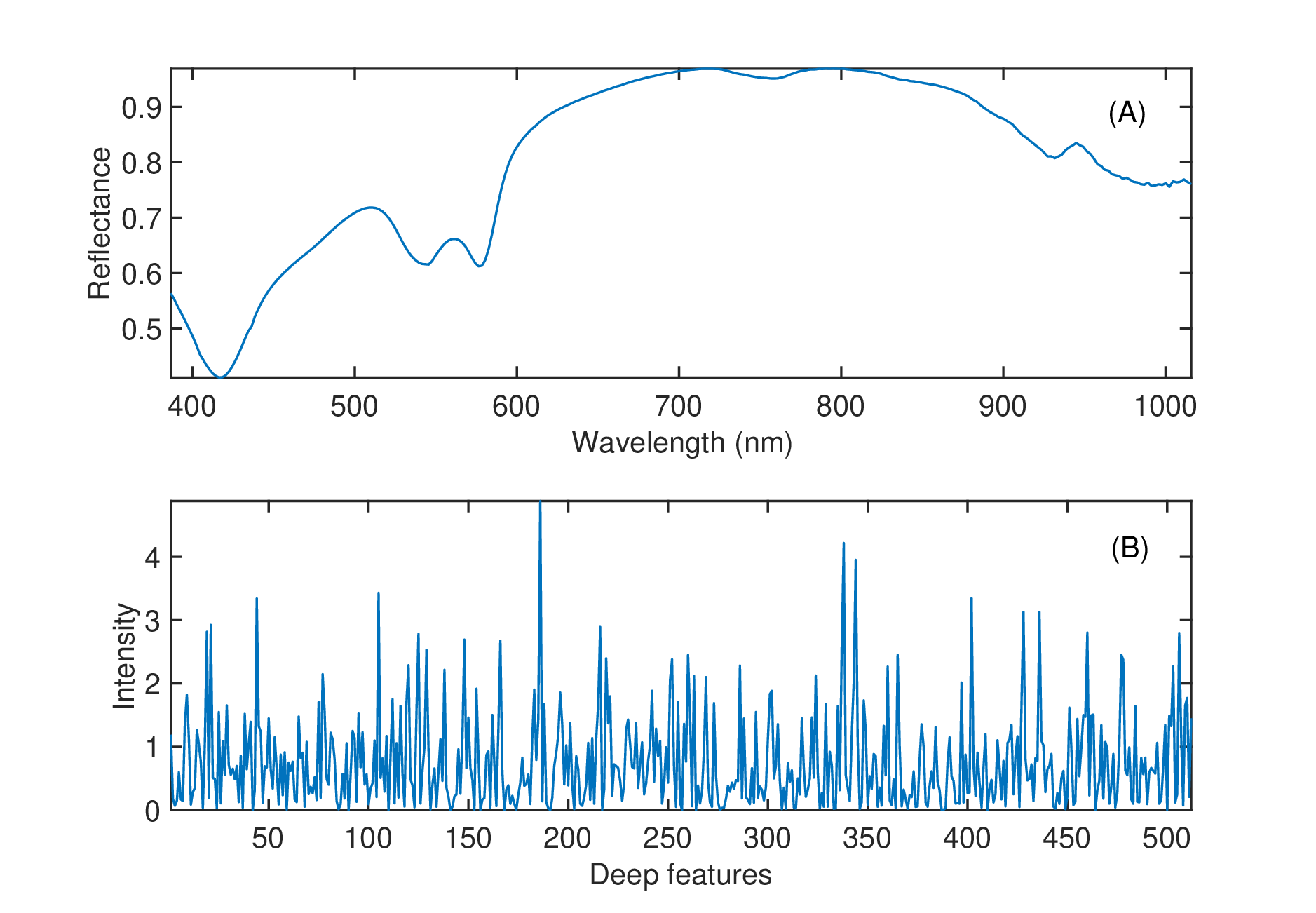}
  \caption{Example coupled mean VISNIR spectra and deep spatial feature vector for single fat sample.}
  \label{fig:fig5}
\end{figure}

The hyperspectral data represent a unique case where information is available from two different modalities, i.e., pseudo RGB images and mean spectra. It is best to use information from both modalities so that any complementary information can be learned and used to explain the response variable. First, the pseudo RGB images can be used to extract deep features using the following code. First, load the deep model:

\begin{lstlisting}[style=Matlab-editor]
%% Sample Matlab code
net = resnet18;
inputSize = net.Layers(1).InputSize;
\end{lstlisting}

Secondly, resizing the images to be used as input to the deep model: 

\begin{lstlisting}[style=Matlab-editor]
%% Sample Matlab code
im = imresize(imread(imagefiles),[224 224]);
augimdsTrain = augmentedImageDatastore(inputSize(1:2),im);
\end{lstlisting}

Thirdly, pass the images through the network to extract deep features as follows:

\begin{lstlisting}[style=Matlab-editor]
%% Sample Matlab code
layer = 'pool5';
featuresTrain = activations(net,augimdsTrain,layer,'OutputAs','rows');
\end{lstlisting}

Once the deep features are extracted, two spectral data matrices are available for predictive modeling: deep features and the mean spectra of the samples. In the chemometrics domain, such a dataset is known as a multiblock dataset, and therefore, multiblock modeling techniques are used for predictive purposes. An example of coupled deep features and mean spectra is shown in Fig. \ref{fig:fig5}. Fig. \ref{fig:fig5} is generated with the following code:

\begin{lstlisting}[style=Matlab-editor]
%% Sample Matlab code
subplot(2,1,1)
plot(wave,reflection;
xlabel('Wavelength (nm)');ylabel('Reflectance');
subplot(2,1,2)
plot(featuresTrain);
xlabel('Deep features');ylabel('Intensity');
\end{lstlisting}

\begin{figure}[H]
  \centering
  \includegraphics[width=\textwidth]{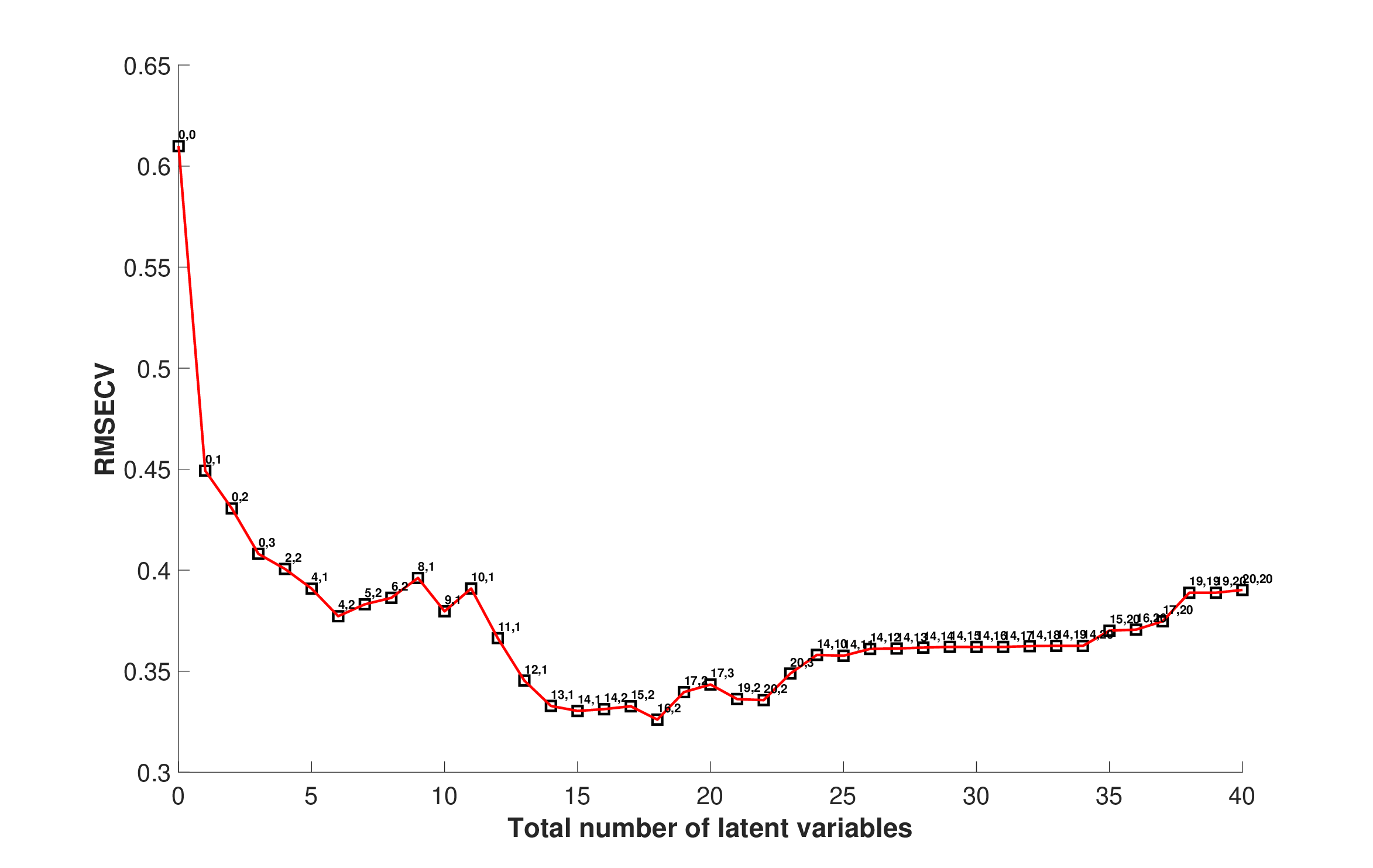}
  \caption{Example coupled mean VISNIR spectra and deep spatial feature vector for single fat sample.}
  \label{fig:fig6}
\end{figure}

A key thing to note is that the scales of the deep features differ from the scale of the reflectance measurements. However, thanks to multiblock methods \cite{mishra2021recent}, the scale difference does not affect the model when information is fused using sequentially orthogonalized predictive modeling. To use both spatial and spectral information, the sequentially orthogonalized PLS \cite{biancolillo2019sequential} regression codes from https://www.chem.uniroma1.it/romechemometrics/research/algorithms/so-pls/ are used. The code is based in MATLAB and can be executed as follows. The code first performs a 5-fold cross-validation analysis to optimize the number of latent variables for each data block. Executing the following code generates a cross-validation curve, as shown in Fig. \ref{fig:fig6}.

\begin{lstlisting}[style=Matlab-editor]
%% Sample Matlab code
model = sopls_cv({spatial_features,spectra_features},y, [20 20], {'mean','mean','mean'});
\end{lstlisting}

\begin{figure}[H]
  \centering
  \includegraphics[width=\textwidth]{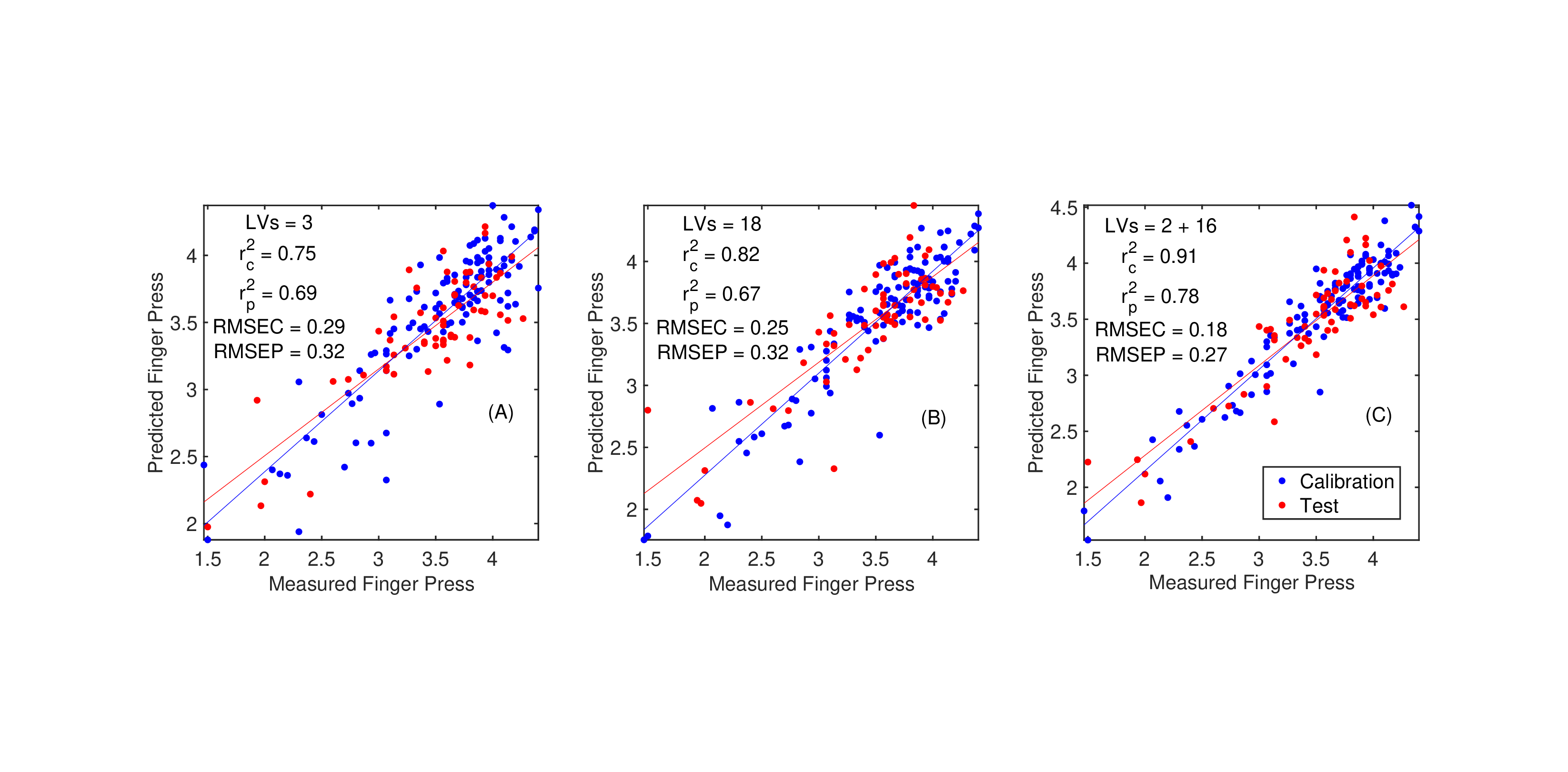}
  \caption{Prediction plot for only spatial (A), only spectral (B) and spatial and spectral fusion (C).}
  \label{fig:fig7}
\end{figure}

Here, the optimal number of components was found to be 2 from spatial information and 16 from the spectral information to predict the response. Having latent variables from both data blocks indicates that the model is learning complementary information from both the spatial and spectral dimensions. The optimal model, based on 2 components from the spatial domain and 16 components from the spectral domain, can be tested. The model test can be performed using the following code, which leads to Fig. \ref{fig:fig7}. In Fig. \ref{fig:fig7}, results from three different models are presented. The first model (Fig. \ref{fig:fig7}A) is purely based on deep features, the second model (Fig. \ref{fig:fig7}B) is purely based on spectral information, and the third model (Fig. \ref{fig:fig7}C) is based on the fusion of spectral and spatial information. As can be seen, the fusion model performed the best, with a prediction error of 0.27, compared to the spatial-only or spectral-only models. These results are in agreement with earlier work \cite{mishra2024short}, where the prediction of a physical property was more accurate when both spatial and spectral information were used.

\begin{lstlisting}[style=Matlab-editor]
%% Sample Matlab code
prediction_fusion = sopls_pred({spatial_features,sp},y, model.OptModel); 
plot(y,prediction_fusion.predY,'.b','MarkerSize',20);
xlabel('Predicted Finger Press');
ylabel('Measured Finger Press');
legend('Calibration','Test');
\end{lstlisting}

\section{Discussion and conclusions}

This tutorial demonstrated the potential of deep feature extraction for solving three distinct analytical problems related to image processing. The accompanying MATLAB codes allow users to perform deep feature extraction across three different imaging modalities: RGB images, X-ray images, and hyperspectral images. Furthermore, it was shown that, in some cases, fusing information from multiple modalities, such as spectral and spatial data, can lead to lower prediction errors. Users are encouraged to run the provided codes on their own image datasets to explore the potential benefits of the method for their specific experiments. The only requirement for potential users is proficiency in MATLAB coding and access to the latest version of MATLAB, for example, 2024 or later. Note that the deep learning model library is not installed by default in MATLAB, and users must install it through MATLAB Add-Ons before attempting this hands-on tutorial.  

In this tutorial, we used ResNet-18 to demonstrate deep feature extraction. However, MATLAB’s model library offers many other pre-trained models, including 'DenseNet201', 'ResNet50', 'ResNet101', 'DarkNet19', 'VGG16', and 'VGG19'. New models are typically added to the library as they are developed in the computer vision domain. If users are proficient in Python, they have access to an even larger selection of up-to-date open-source models, such as the pre-trained models based on the transformer architecture\cite{dosovitskiy2021imageworth16x16words}. Different models capture varying levels of abstraction, and choosing the best model for a particular application may require some research into the type of dataset the model was trained on and the function of specific layers in its architecture. Alternatively, users could select newer models that have typically outperformed older ones. However, some models are very large and computationally intensive, so users should also consider the computational requirements. Information on computation times is often included in the scientific articles where these models were first introduced. A potential direction for future research is to extract features from multiple pre-trained deep models and fuse them using multi-block data modeling techniques \cite{mishra2021recent}.  

Deep learning models are sequential in nature, with different layers that capture information at varying levels of abstraction. This sequential nature allows information to be extracted from any step in the model. The most common practice is to extract features from the second-to-last layer, as it provides the most condensed information with multiple levels of abstraction. However, users are free to experiment with feature extraction from different layers of the model. A potential advantage of using fewer layers for feature extraction is reduced computation time, as fewer layers require less processing. To extract features from different layers, users need to modify the 'layer' parameter. For instance, in this tutorial, the 'layer' parameter was set to 'pool5', meaning that features were extracted from the 5th global pooling layer. To extract features from the fourth global pooling layer, users would need to specify `layer = 'pool4'`. To view the names of all layers available in the model, users can open the model in the Model Designer, which can be accessed by double-clicking the loaded model in the MATLAB workspace.  

Despite their advantages, pre-trained deep learning models present several challenges. They are computationally demanding, especially when applied to high-resolution images or large datasets. Computational limitations depend on both the hardware used and the model architecture selected. For example, in applications requiring real-time, in-line processing, deep feature extraction may introduce computational latencies, depending on the speed requirements of the system. Another drawback is that the features extracted by deep networks are often abstract and may lack interpretability, posing a challenge to chemists and other domain scientists who need clear and actionable insights. For example, features extracted from different datasets appear as collections of peaks, offering limited interpretability. However, these features carry valuable information for exploratory and predictive modeling. Linking extracted features to meaningful interpretations remains an open area of research, which could further complement their predictive potential. This study offers the MATLAB tutorial to reduce the scripting burden for deep feature extraction for domain scientists, bridging the gap between computer vision applications and analytical chemistry.

In the presented tutorial, only deep learning models that accept RGB images as input were explored. Therefore, all images with more than three spectral dimensions must first be compressed to three dimensions. However, a wide range of deep learning models are available for processing different types of inputs. For example, there are models capable of extracting features from one-dimensional signals, 3D images, and even videos. These models can be selected based on the type of data being processed. If users are uncertain about which network to use, generative models, such as ChatGPT and Perplexity, can assist in identifying the most suitable and up-to-date models.  

This study also presented an example of spatial-spectral fusion. In the field of spectral imaging processing, data analysis is typically limited to utilizing only the spectral information from the imaged samples, with spatial information being rarely used. However, recent studies have shown that combining spatial and spectral information can be beneficial, often leading to improved predictions of properties of interest compared to using either spatial or spectral data alone. Such fusion, however, must be performed properly to model only the complementary information from the spatial and spectral domains. Multi-block methods in the domain of chemometrics allow for such types of modeling \cite{mishra2021recent}.  Recent work has demonstrated that fusing spectral and spatial information \cite{mishra2024short} is particularly beneficial for parameters related to both the physical and chemical properties of samples. However, for parameters that are purely chemical, the inclusion of spatial information does not offer significant benefits. In such cases, spectral information alone is sufficient for accurate predictions.

Pre-trained deep learning models, in addition to feature extraction, can be used to automate several other image processing tasks in the analytical chemistry domain. For instance, recent object detection and segmentation models, such as the Segment Anything Model or YOLO, can be directly applied to images to perform real-time object detection, localization, and segmentation tasks. If any adaptation is needed, the pre-trained model can also be fine-tuned using transfer learning. Some recent examples of using pre-trained models for object detection can be found in \cite{xu2022combining,xu2022complementary,puneet2022deep}.

\section{Acknowledgments}

This research is supported by the KB-54 Sustainable Nutrition $\&$ Health within Wageningen University $\&$ Research and received financing from the Dutch ministry of Agriculture, Fisheries, Food Security and Nature.\\
CHATGPT for English grammar correction.\\
Project RTI2018-096993-B-I00 financed by Spanish Ministry of Science, Innovation and Universities and FEDER.

\printbibliography

\end{document}